\documentclass[runningheads]{llncs}
\usepackage{graphicx}
\usepackage{comment}
\usepackage{amsmath,amssymb} 
\usepackage{color}

\usepackage[width=122mm,left=12mm,paperwidth=146mm,height=193mm,top=12mm,paperheight=217mm]{geometry}
\usepackage{booktabs}
\usepackage{tabto}
\usepackage{tabu}

\usepackage{mathrsfs}
\usepackage{dsfont}
\usepackage{multirow}
\usepackage{soul}
\usepackage[table]{xcolor}
\usepackage{subfig} 
\usepackage{gensymb}
\usepackage{stackengine}

\definecolor{lightgrey}{rgb}{0.925, 0.925, 0.925}
\sethlcolor{lightgrey}

\definecolor{lightblue}{rgb}{0.4, 0.8, 0.99}
\sethlcolor{lightblue}

\definecolor{TableRed}{HTML}{800000}
\newcommand{\TextRed}[1]{\textcolor{TableRed}{#1}}

\newcommand{\Est}[1]{\hat{#1}}
\DeclareMathOperator*{\argmax}{arg\,max}
\newcommand{\DATASETNAME}{Surface Property Synesthesia}
\newcolumntype{?}{!{\vrule width 1pt}}

\begin{document}

\pagestyle{headings}
\mainmatter

\title{Teaching Cameras to Feel: Estimating Tactile Physical Properties of Surfaces From Images} 

\titlerunning{Teaching Cameras to Feel}
%
\author{Matthew Purri \and
Kristin Dana}
%
%
\institute{Rutgers University}
\maketitle

\begin{abstract}
\vspace{-0.3cm}
The connection between visual input and tactile sensing is critical for object manipulation tasks such as grasping and pushing.
In this work, we introduce the challenging task of estimating a set of tactile physical properties from visual information. We aim to build a model that learns the complex mapping between visual information and tactile physical properties. We construct a first of its kind image-tactile dataset with over 400 multiview image sequences and the corresponding tactile properties. A total of fifteen tactile physical properties across categories including friction, compliance, adhesion, texture, and thermal conductance are measured and then estimated by our models. We develop a cross-modal framework comprised of an adversarial objective and a novel visuo-tactile joint classification loss.
Additionally, we introduce a neural architecture search framework capable of  selecting optimal combinations of viewing angles for estimating a given physical property.

\vspace{-0.05cm}
\end{abstract}

\section{Introduction}
\begin{figure}
    \centering
    \subfloat{\includegraphics[width=0.25\textwidth]{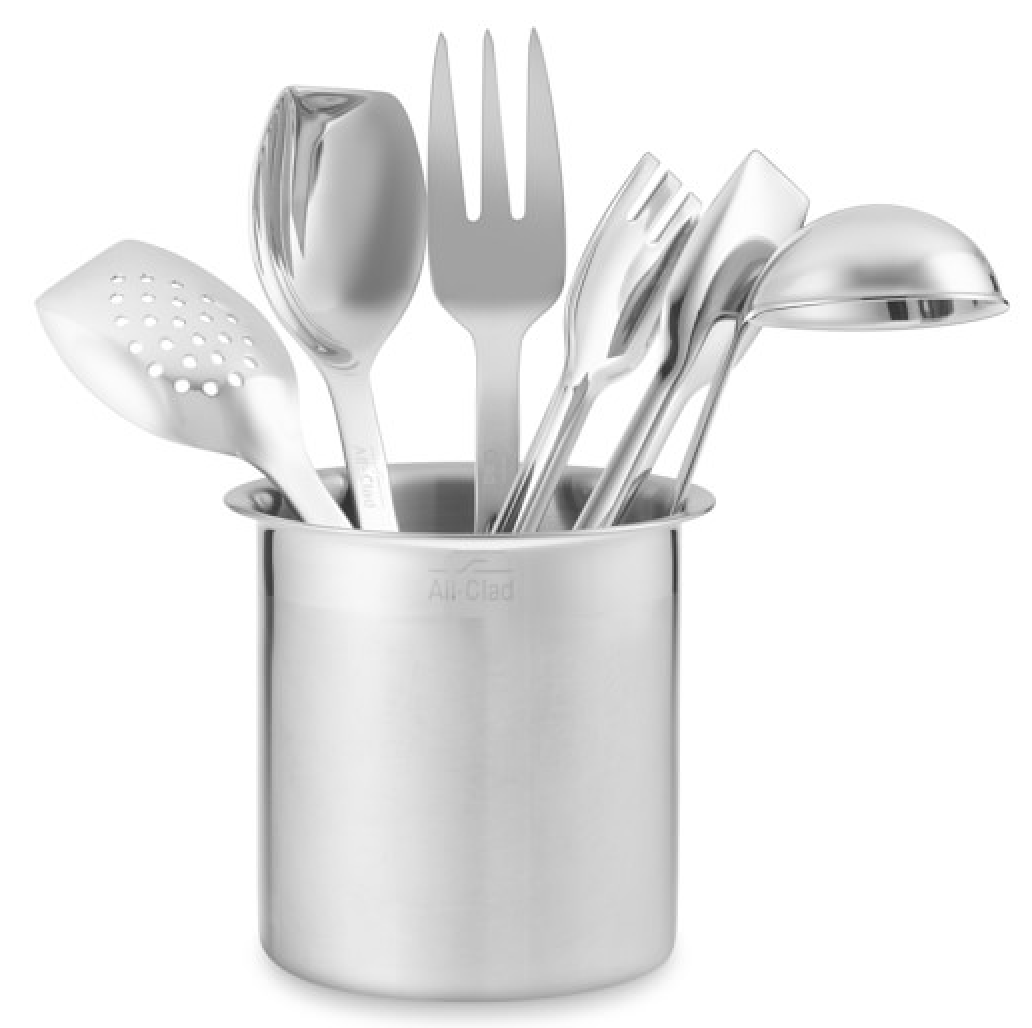}} \hspace{0.1cm}%
    \subfloat{\includegraphics[width=0.70\textwidth]{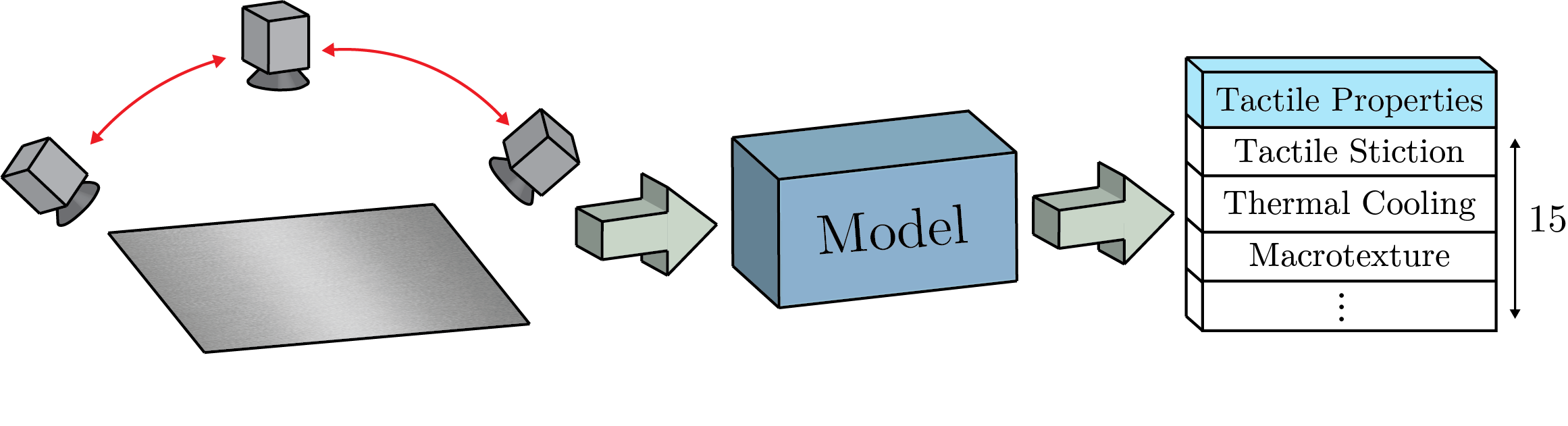}}
    \caption{\small{\textbf{Material example and inference framework.}  \textbf{Left:} An example of objects with different geometry and semantic labels that share common physical properties. \textbf{Right:} The proposed inference model receives images taken at various viewing angles and predicts a set of tactile property values.
    }} \vspace{-0.4cm}
    \label{fig:main_figure}
\end{figure}

In real-world tasks such as grasp planning and object manipulation, 
humans infer physical properties of objects from visual appearance.
%
Inference of surface properties is distinct from object recognition. For example in Figure~\ref{fig:main_figure}a, the objects have different geometric shape; however, they share similar tactile physical properties. 
We can imagine what it would feel like to pick one up and handle it.
Recognition can provide the semantic labels of the utensils, but tactile inference  can provide  the physical properties of the stainless steel.
In this work, we introduce a computational model that learns the complex relationship between visual perception and the direct tactile physical properties of surfaces such as compliance, roughness, friction, stiction, and adhesive tack.  

There are many instances where an accurate estimate of a surface’s tactile properties is  beneficial for automated systems. In colder climates for example, thin layers of ice form over driving surfaces dramatically decreasing the sliding friction of a road. Modern vision systems trained on autonomous driving datasets such as KITTI~\cite{geiger2013vision} or Cityscapes~\cite{cordts2016cityscapes}  can readily identify ``road'' pixels, but would not provide the coefficient of friction required for braking control. 
Another example is manufacturing garments or shoes that require precise manipulation of multiple types of materials. Delicate and smooth fabrics such as silk require different handling than durable denim fabric. 
Also, robotic grasping and pushing of objects in a warehouse can benefit from   surface property estimation to improve manipulation robustness.   

\vspace{-0.20cm}

In recent years, there has been increased interest in estimating physical properties of objects and forming learned physics engine models to infer the reaction of an object to controlled stimuli. Many of these methods passively observe interacting objects or select actions to stimulate object movement, learning to map visual information to physical properties by estimating the effect of the action on the scene ~\cite{wu2015galileo,wu2017learning,xu2019densephysnet,janner2018reasoning}. 
In these methods, geometric and physical properties of the objects are encoded into latent space, and  
a learned physics engine predicts the future state of the objects.
The indirect representation of the object properties confounds high-level actions,
such as pushing with the precise amount of force to overcome friction or ordering surfaces based on their roughness. 
In our work we estimate the physical properties directly, allowing attributes of objects to be utilized directly. 

We formulate the challenge of estimating a surface's physical properties from visual information as a cross-modal translation problem. Cross-modal problems remain one of the frontiers in vision research. For example, many recent works relate disparate information streams such as videos with audio~\cite{owens2018audio,owens2016ambient,arandjelovic2018objects,owens2016visually} and images with captions~\cite{nam2017dual,karpathy2015deep,kiros2014unifying}.
Here, the problem of translating images into tactile properties provides a unique and challenging task. The visual and tactile modalities are not aligned in the same manner as audio and video  and the scale discrepancy between images and tactile property vectors is vast.

To address this challenge, we create a dataset of 400+ surface image sequences and 
tactile property measurements. 
Given that humans have the ability to estimate tactile properties based on experience and visual cues, we expect autonomous systems to also be able to learn this complex mapping. While estimating surface properties, people often unconsciously move their heads,  acquiring multiple views of a surface. Inspired by this, the  captured images sequences comprise multiple viewing angles for each material surface.

Some challenges can be solved effectively using a single view of a scene, such as object classification, semantic segmentation, or image denoising. Whereas tasks such as 3D geometry reconstruction and action recognition require more information than what a single image can typically provide. A rich literature of material recognition, a similar challenge to surface property estimation, has found advantages to using multiple viewpoints or illumination conditions to identify the material class of a surface. For example, reflectance disks~\cite{zhang2015reflectance,zhang16}, optimal BRDF sampling~\cite{jehle2010learning,liu2014discriminative,nielsen2015optimal}, angular gradient images~\cite{xue2017differential}, 3D point clouds~\cite{degol2016geometry,zhao2017fully}, 4D light field images~\cite{wang20164d}, and BRDF slices~\cite{wang2009material} all provide good material recognition performance with partial viewpoint or illumination sampling. These methods, however, rely on sampling at a fixed set of viewing angles. 
In this work, we allow our model to select the optimal partial BRDF for tactile property estimation, providing empirical insight for which viewpoints should be sampled in camera motion planning and physical property sensor designs.

The main objective of this work is to leverage the relation between the appearance of material surfaces and their tactile properties to create a network capable of mapping visual information to tactile physical properties.
We have three main contributions. 
First, we develop a visual-tactile dataset named  
\DATASETNAME~dataset; with 400+ material surfaces imaged at 100 viewing angles and augmented with fifteen measured tactile physical properties (as listed in Table 1) measured by a BioTac Toccare tactile sensor. \footnote{Tactile measurements were done by SynTouch Inc., with the BioTac Toccare, and purchased by Rutgers.} Second, we propose a cross-modal learning framework with adversarial learning and cross-domain joint classification to estimate tactile physical properties from a single image. Third, we introduce two input information sampling frameworks that learn to select viewing angle combinations that optimize a given objective.
Our results show that image-to-tactile estimation is challenging, but  we have made a pioneering step toward direct tactile property estimation.

\vspace{-0.15cm}

\section{Related Work}
\vspace{-0.05cm}
\subsubsection{Cross-modal Learning and Translation} 

Cross-modal translation is defined as finding a function that maps information from one modality into its corresponding representation in a different modality. Prototypical approaches involve embedding data from different modalities into a learned space, from which generative functions map embeddings to their original representations. Recent works have applied this framework to various modality combinations including image-to-text~\cite{salvador2017learning,zhang2018deep,karpathy2015deep,ranjan2015multi,wang2017adversarial}, image-to-image~\cite{yuan2017connecting,zhang2018translating,zhu2017toward}, audio-to-image~\cite{taylor2017deep,chen2017deep}, and audio-to-text~\cite{chung2018unsupervised}. Aytar et al. create a text, audio, and image multimodal retrieval system by leveraging large unaligned data sources~\cite{aytar2017see}. Example-based translation methods such as retrieval systems rely on a dictionary (training data) when translating between modalities, whereas generative translation models directly produce translations. Generative translation~\cite{gonzalez2018image,li2019connecting} is considered a more challenging problem than example-based translation because a generative function must be learned in addition to embedding functions. In~\cite{gonzalez2018image}, shared and domain-specific features are disentangled through auxiliary objectives to produce more realistic image-to-image translations. Generative translation models require large combinations of modality pairs to form a representative latent space. In this work, we introduce a new dataset containing a novel modality combination of image sequences and tactile physical properties. 

\vspace{-0.3cm}

\subsubsection{Visuo-Tactile} 

There is much interest in both the vision and robotics communities in giving robots the ability to understand the relation between visual and haptic information. A variety of challenges have been solved more efficiently with the addition of tactile information including object recognition~\cite{falco2017cross}, material classification~\cite{kerzel2017haptic}, and haptic property estimation~\cite{gao2016deep}. Calendra et al. combine a GelSight sensor~\cite{li2014localization,yuan2017connecting,li2019connecting,calandra2018more,yuan2018active} with an RGB camera to jointly predict grasp outcomes and plan action sequences for grasping~\cite{calandra2018more}. Gao et al. improve the performance of a haptic adjective assignment task by fusing images and time sequence haptic measurements~\cite{gao2016deep}. The aim of this work is not to improve the performance of a particular task but to find the relationship between visual information and touch, such that tactile properties can be estimated from visual information. Recently, works such as~\cite{yuan2017connecting} and~\cite{li2019connecting} similarly seek to learn a mapping between vision and touch either by learning a latent space for retrieval or by synthesizing realistic tactile signals from visual inputs. In contrast to these works, we directly generate tactile estimates and the representation of our tactile representation is a physical property vector instead of a tactile image. Most similar to our work, Zhang et al. estimate the coefficient of friction of a surface from a reflectance image~\cite{zhang2016friction}. We expand upon previous work to estimate a larger set of fifteen tactile properties including friction, texture, thermal conductance, compliance, and adhesion. Additionally, we assume that the visual information to our system consists of ordinary images obtained by standard RGB camera sensors.

\vspace{-0.3cm}

\subsubsection{Viewpoint Selection} 
Given an oversampled set of images, how can we select the most useful subset of images from that set? 
In this work, we capture an oversampled sequence of images measured at consistent viewpoints via a gonioreflectometer. Inspired by viewpoint and illumination selection techniques for BRDFs~\cite{nielsen2015optimal,jehle2010learning,liu2014discriminative,xue2017differential}, our aim is to determine what combination of viewing angles produce the optimal output for physical property estimation. Nielsen et al. determine the minimum number of samples to reconstruct a measured BRDF by randomly sampling viewing and illumination angles and comparing their condition numbers~\cite{nielsen2015optimal}.
Xue et al. capture pairs of images with small angular variations to generate improved angular gradients which serve as additional input for material recognition networks~\cite{xue2017differential}. In contrast to viewpoint trajectory optimization~\cite{johns2016pairwise,wu20153d,jayaraman2018learning}, where the objective is to actively plan a viewing path in SO(3) space, to decrease task uncertainty,  our objective is specifically to improve image-to-tactile estimation performance. 
In video understanding tasks, selectively sampling frames can enable efficient pattern analysis~\cite{zhou2018temporal,feichtenhofer2019slowfast,shelhamer2016clockwork}. 
In~\cite{zhou2018temporal}, features from random subsets of frames of video are extracted and summed into a final representation to efficiently improve action recognition.  Similarly, we seek to  sample the angular space of multiview images. However, our approach learns the optimal sampling for the task at hand. 


Inspired by neural architecture search (NAS) approaches~\cite{pham2018efficient,liu2018progressive,zoph2016neural,liu2018darts}, we learn to select a combination of viewing angles instead of choosing a handcrafted selection strategy similar to Xue et al.~\cite{xue2017differential}. NAS methods are comprised of a search space, search strategy, and a performance estimation strategy. Generally, the search space of NAS is a set containing all possible layer functions and layer connections, whereas the search space for our problem are all combinations of viewing angles. To our knowledge, we are the first to utilize NAS for searching over the input space, instead of the overall architecture, resulting in a viewpoint selection.  
We propose two NAS frameworks for learning combinations of viewing angles which improve tactile physical property estimation as well as providing insight into what combinations of viewpoints are most informative for estimating a given physical property.

\vspace{-0.6cm}

\begin{table}[]
\centering
\caption{\small{\textbf{Tactile property acronyms.} Acronyms for each of the fifteen tactile properties measured by the Toccare device.}} \vspace{0.1cm}
\resizebox{0.98\columnwidth}{!}{%
\begin{tabular}{|c|c|c|c|c|c|}
\hline
fRS & Sliding Resistance     & fST & Tactile Stiction        & uCO & Microtexture Courseness \\ \hline
uRO & Microtexture Roughness & mRG & Macrotexture Regularity & mCO & Macrotexture Courseness \\ \hline
mTX & Macrotexture           & tCO & Thermal Cooling         & tPR & Thermal Persistance     \\ \hline
cCM & Tactile Compliance     & cDF & Local Deformation       & cDP & Damping                 \\ \hline
cRX & Relaxation             & cYD & Yielding                & aTK & Adhesive Tack           \\ \hline
\end{tabular}
}
\label{tab:acronyms}
\end{table}

\vspace{-0.8cm}

\section{\DATASETNAME~Dataset}
\label{section:dataset}

\begin{figure}
    \centering
    \subfloat{\includegraphics[width=0.32\linewidth]{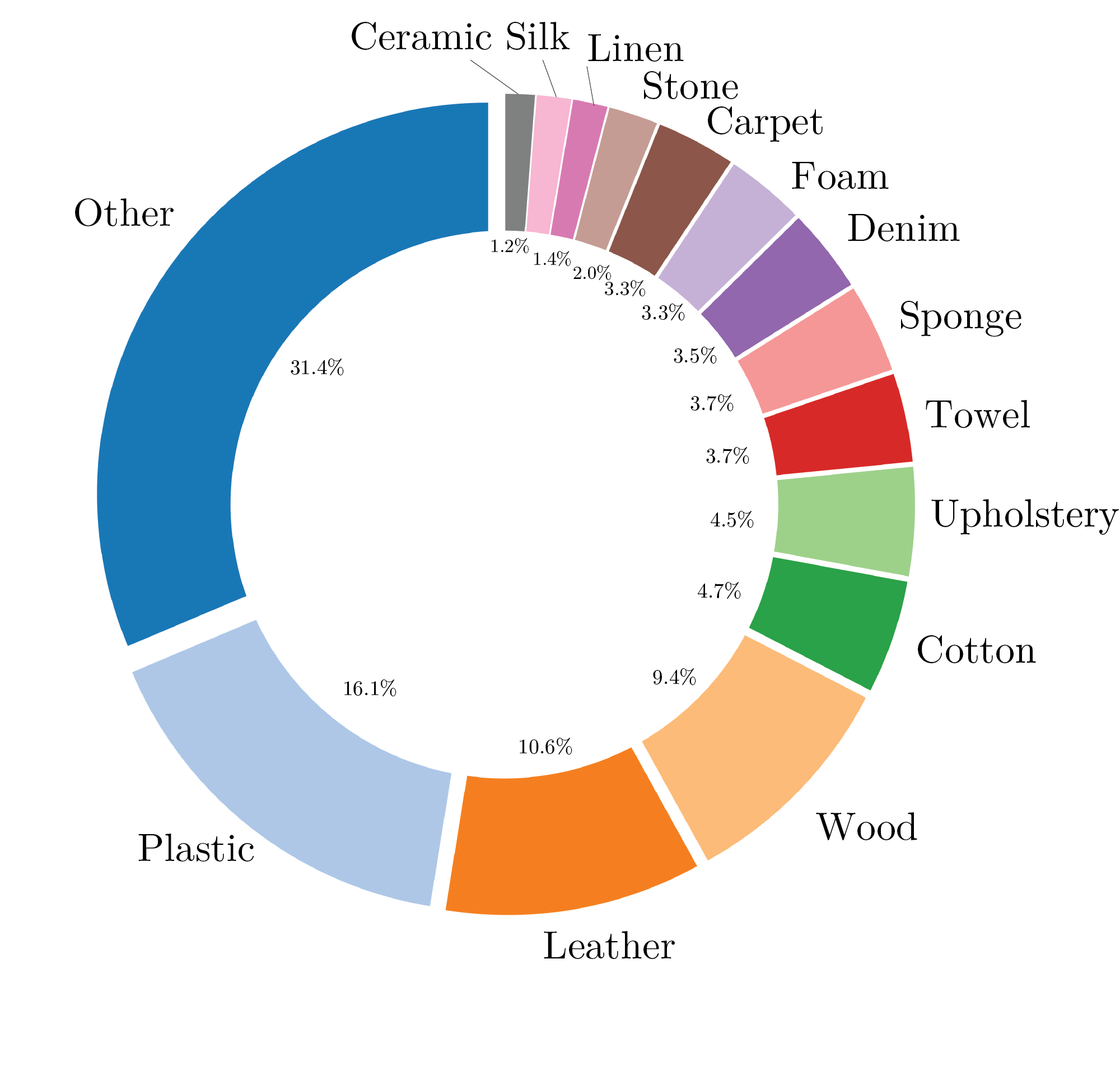}} \hspace{0.1cm}%
    \subfloat{\raisebox{0.5cm}{\includegraphics[width=0.33\linewidth, height=0.25\linewidth]{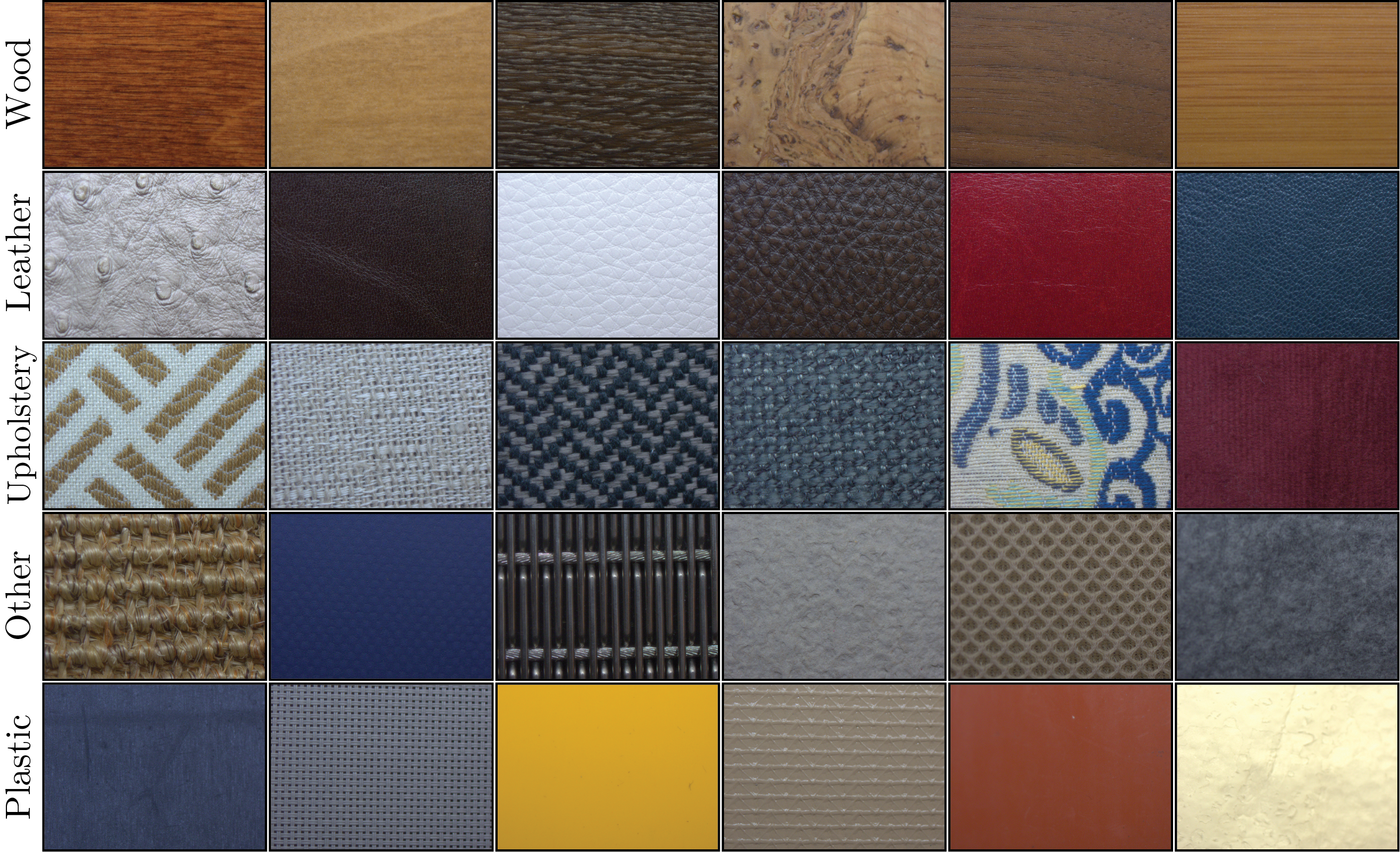}}} \hspace{0.07cm}%
    \subfloat{\includegraphics[width=0.32\linewidth, height=0.32\linewidth]{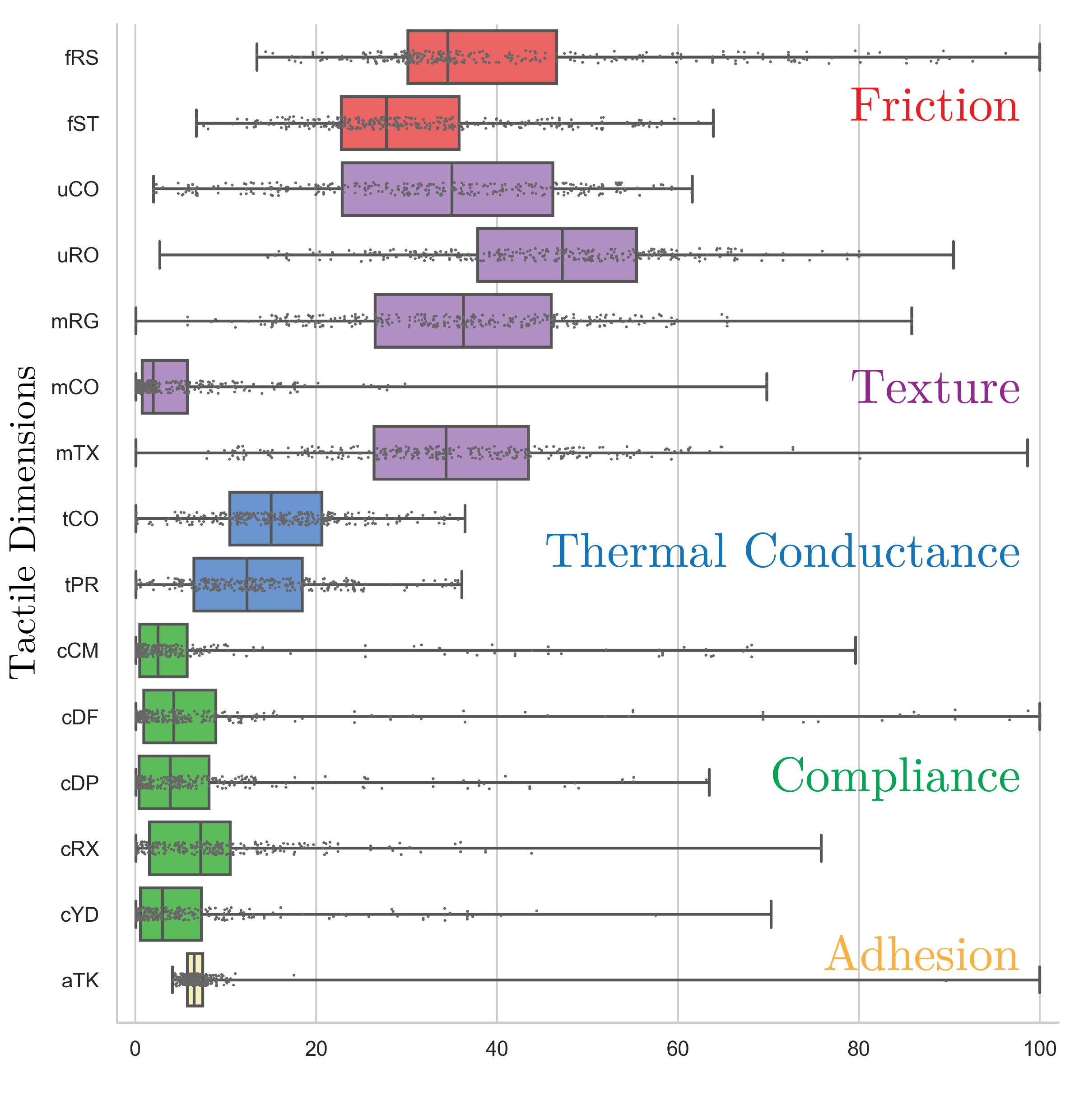}}
    \caption{\small{\textbf{Dataset statistics.} \textbf{Left:} The distribution of material labels. The output of the models are tactile properties and not material labels. The other category includes materials with less than six occurrences. \textbf{Center:} A sampling of the 400+ materials in our dataset, which highlight the diversity in visual appearance. The other category includes materials such as metal, fur, corduroy, nylon, and more. \textbf{Right:} A box plot distribution for each tactile property. The acronym for each property is defined in Table~\ref{tab:acronyms}.}}
    \label{fig:dataset_stats}
\end{figure}

Synesthesia is the production of an experience relating to one sense by a stimulation of another sense. For example, when viewing an image of a hamburger you may unconsciously imagine the taste of the sandwich. In this work, images of surfaces are perceived and the tactile properties of that surface are estimated. To train a model for tactile physical property estimation, we collect a dataset named the {\it Surface Property Synesthesia Dataset} (SPS) consisting of pairs of RGB image sequences and tactile measurements. The dataset contains 400+ commonly found indoor material surfaces, including categories such as plastic, leather, wood, denim, and more as shown in Figure~\ref{fig:dataset_stats}a. To our knowledge, this dataset contains the largest number of material surfaces of any visuo-tactile dataset, a necessity for learning the complex relation between vision and touch. A majority of the dataset belongs to four of the fifteen material categories. However, each category contains a diverse set of surfaces in terms of both color and pattern as shown in Figure~\ref{fig:dataset_stats}b. \textit{The dataset and source code are made publicly available.}\footnote{\url{https://github.com/matthewpurri/Teaching-Cameras-to-Feel}}

\subsubsection{Tactile Data}

The Biotac Toccare is a tactile sensing device that measures fifteen tactile physical properties of a surface and has been shown to identify materials more accurately than people~\cite{fishel2012bayesian}. The device is comprised of a BioTac sensor~\cite{su2015force,arian2014using,kerr2018material,reinecke2014experimental} used to collect a series of time-varying signals and a staging system to consistently move the tactile sensor over a surface. Low-frequency fluid pressure, high-frequency fluid vibrations, core temperature change, and nineteen electrical impedances distributed along the sensor surface are recorded over time as the sensor is in contact with a surface. The signals are then converted into fifteen tactile physical properties whose values range from 0 to 100. Surface property measurements are gathered across several locations on the surface. Each tactile measurement is repeated five times.

The fifteen physical properties used to describe a surface, can be organized into five major tactile categories including friction, texture, thermal conductance, compliance, and adhesion as shown in Figure~\ref{fig:dataset_stats}c. Specific descriptions for each of the fifteen properties are described in the supplementary material. The \textit{texture} category represents both macro and micro-texture surface attributes which correspond to large and small height intensities along a surface. Both static and kinetic friction are included in the \textit{friction} class. \textit{Adhesion} describes the perceived effort to break contact with a surface with values semantically ranging from no adhesion to sticky. The rate of heat transferred from the BioTac sensor to the surface is described in the \textit{thermal conductance} category. Surface deformation characteristics correspond to the \textit{compliance} category.

\subsubsection{Vision Data}

After tactile property measurements are obtained for each surface, images of the same surfaces are taken with a gonioreflectometer.  The surfaces are imaged in a continuous manner from $-45\degree$ to $45\degree$ along the roll axis of the surface. For each material surface, 100 images are recorded. The yaw and pitch angles are constant throughout the imaging sequence. All images were taken under a mostly diffuse light source. 

\begin{figure}[!ht]
    \centering
    \includegraphics[width=0.90\columnwidth]{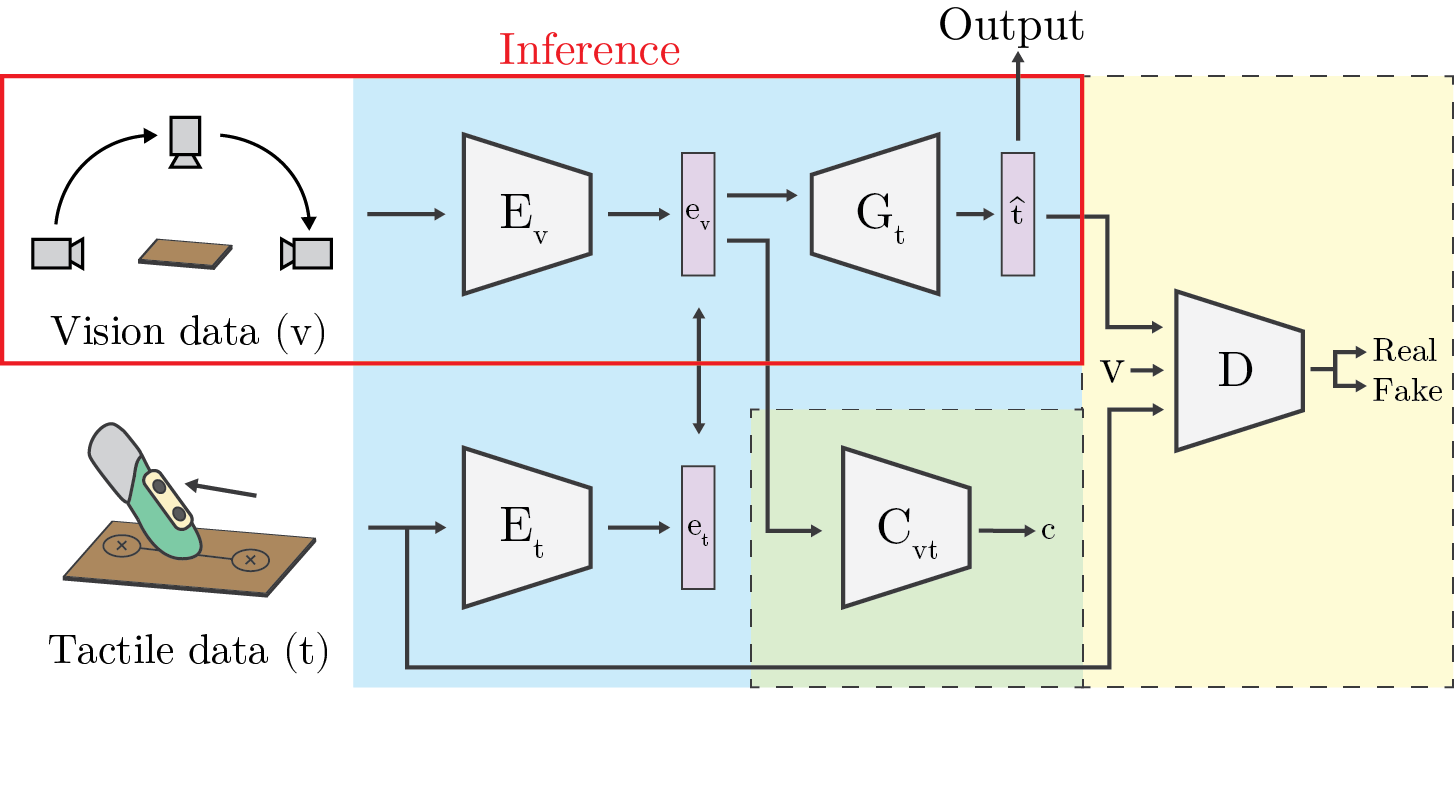}
    \vspace{-0.75cm}
    \caption{\small{\textbf{Overview of our proposed cross-modal framework.} The model is comprised of four modules: latent space encoding (blue), feature-classification (green), adversarial learning (yellow), and viewpoint selection (not displayed). The objective of this model is to generate precise tactile properties estimates $\Est{\mathrm{t}}_i$ given vision information $v_i$. Both visual and tactile information (measured with Toccare device) are embedded into a shared latent space through separate encoder networks and compared.
    A generator function $\mathrm{G_t}$ estimates tactile property values $\Est{\mathrm{t}}$ from the embedded visual vector $\mathrm{e_v}$. The discriminator network $\mathrm{D}$ learns to predict whether a tactile-visual pair is a real or synthetic example. An auxiliary classification network $\mathrm{C_{vt}}$ generates a visuo-tactile label given $\mathrm{e_v}$. The modules included in the red boundary represent the networks used during inference. Note, no tactile information is used during inference.}}
    \label{fig:cross-modal_arch}
\end{figure}

\section{Methods}
\label{sec:methods}


\subsection{Mapping Vision to Touch}
\label{section:method_single_image}

\subsubsection{Problem Definition}
We model the problem of translation between two modalities as a cross-modal translation problem. We are specifically interested in the translation from images to tactile properties. Let $t_i \in \mathbb{R}^{D}$ represent tactile physical property vectors and $v_i \in \mathbb{R}^{3 \times F \times H \times W}$ represent image sequences where $F$, $H$, and $W$ correspond to the number of frames, height, and width respectively. Instances of each modality are encoded through distinct encoding networks, $E_t$ and $E_v$, into a shared latent space. In this space, embedded visuo-tactile pairs $e^t_i$ and $e^v_i$ should be close and are encouraged to be near each other by a pairwise constraint $\mathcal{L}_{emb} = \Vert e^t_i - e^v_i \Vert_2^2$. Estimated tactile physical property vectors are created via a generative function $G_t$, given the embedded representation of the visual information as input $G_t(e^v_i) = \Est{t}_i$.

\subsubsection{Regression Baseline}
To evaluate the capabilities of the cross-modal network, we compare its results to the results obtained from a regression network. The regression network encodes a single image of a material surface into a tactile physical property estimate omitting the intermediate latent representation and embedding constraint, $E_t(v_i) = \Est{t}_i$. 

\subsubsection{Cross-Modal Network}

\vspace{-0.2cm}

\paragraph{Adversarial Objective}

Inspired by multi-modal works~\cite{li2019connecting,zhang2018translating,zhu2017toward}, we augment our cross-modal framework with an adversarial objective in order to improve the quality of the tactile estimates. The input visual information is combined with the estimated tactile information and then used as input into a discriminator network to determine if the pair is real or fake. This process forms the following objective:

\vspace{-0.5cm}

\begin{equation}
\begin{split}
	\mathcal{L}_{adv}(G_t, D) = &\mathbb{E}_{v,t} [ \log D( v, t ) ] + \\ &\mathbb{E}_{v,t}[ \log (1 - D(v, G_t(e_v)) ] + \mathbb{E}_{v,t} [\Vert G_t(e_v) - t \Vert_2],
\end{split} 
\end{equation}
where the generator $G_t$ attempts to generate realistic tactile property vectors that are conditioned on the embedded visual information $e_v$ while the discriminator $D$ tries to distinguish between real versus fake visuo-tactile pairs. In prior work on conditional GANs~\cite{isola2017image,dai2017towards,perarnau2016invertible}, the input and output of the generator are the same dimension whereas the input of our generator can be a sequence of images and the output is a low dimensional vector. In order to handle the scale difference, we combine the tactile property estimation with the feature vector output of a single image instead of the full resolution image. The feature vector is generated via the image encoding network $E_v$.


\paragraph{Classification Objective}
A key to forming a latent space that is well conditioned for a given objective is to add constraints to that space. In addition to constraining each visuo-tactile embedding pair to be close to each other in latent space, it would be advantageous for surfaces with similar physical properties to be close. Other works~\cite{salvador2017learning,zhang2018deep} have included this clustering constraint by adding an auxiliary classification objective. For many problems, semantic labels are informative of the properties that objects contain, however in our case the material labels are not always informative of tactile properties, e.g. plastics come in many forms and possess distinct surface properties but fall under one label. Yuan et al. circumvent this challenge by creating pseudo-labels formed through clustering hand-labeled tactile properties~\cite{yuan2017connecting}. We extend unsupervised cluster labeling by creating labels from visuo-tactile embedding clusters instead of only tactile property clusters. 
Examples with similar tactile properties and visual statistics are encouraged to be close in space by this objective.
Visuo-tactile representations are generated first by encoding features of one of the images in the sequence with a model pretrained on ImageNet~\cite{russakovsky2015imagenet}. The dimensionality of the feature vector is reduced through PCA and normalized to zero mean and unit variance. The reduced visual feature vector and tactile property vector are concatenated to form the final visuo-tactile representation. K-means is then used to cluster the visuo-tactile representation, creating $k$ labels.

The adversarial and classification auxiliary objectives are combined with the tactile property estimation loss and cross-modal embedding constraint to form the final cross-modal objective:

\begin{equation}
    \mathcal{L} = \mathcal{L}_{est} + \lambda_1                              \mathcal{L}_{emb} + \lambda_2                              
    \mathcal{L}_{adv} + \lambda_3
    \mathcal{L}_{class}.
\end{equation}

\subsubsection{Evaluation Metric}
We use the coefficient of determination ($\mathcal{R}^2$),  mean absolute error (MAE), and median percentage error ($\%_{err}$) metrics to evaluate how close each estimated tactile vector is to the ground truth values. The top eight percentage error ($\%_{err}^{T8}$) is used to compare network performances on the eight best performing tactile properties in terms of $\%_{err}$. The top eight tactile properties are selected based on the metrics shown in  Table~\ref{tab:single-image-tactile-reconstruction_perr}. The $\mathcal{R}^2$ metric has been used to access the performance of vision-based regression tasks in several works~\cite{glasner2015hot,volokitin2016deep,bessinger2016quantifying,mccurrie2017predicting,jean2016combining,burgos2014distance,dymczyk2016erasing}. The $\mathcal{R}^2$ metric compares the estimation from the model $\Est{t}$ against using the mean tactile value $\overline{t}$ from the training set as an estimation, and is given by:

\vspace{-0.2cm}

\begin{equation}
    \mathcal{R}^2 = 1 - \frac{\sum_{i} (t_i - \Est{t}_i)^2}{\sum_{i} (t_i - \overline{t})^2}.
\end{equation}

\vspace{-0.1cm}

\subsection{Viewpoint Selection}
\label{section:method_multi_image}

\subsubsection{Viewpoint Selector Framework}

As mentioned in Section~\ref{section:dataset}, each material surface in the 
SPS dataset is oversampled in viewing angle space. 
Selectively sampling viewing angles has been shown to improve performance for action classification tasks over using all available information~\cite{zhou2018temporal}. The challenge of selectively sampling viewing angles is formulated as follows: given a set $p$ of $N$ images collected from distinct viewing angles, select an optimal combination $q$ of $M$ images that minimize the tactile estimation error, $q^\star = \{ \min \Vert t - f(q;w_\theta) \Vert_2^2 \mid q \subset p, \vert q \vert = M , \vert p \vert = N\}$. The combinatorics of this problem are too vast to explore fully, therefore we construct a sampling network $\pi(w_{\pi})$ tasked with learning to select the optimal combination of viewing angles $q^\star$ based on weights $w_{\pi}$. The optimal combination is then used as input for a tactile estimation network $f(q^\star;w_\theta)$. We call the sampling network the Neural Viewpoint Selector (NVS) network. The NVS network is comprised of $M$ viewpoint selector vectors $z$, each tasked with choosing a single viewpoint from $N$ possible viewpoints. Each viewpoint image is assigned an equal probability of being selected. The NVS module selects a single viewing angle for the set $q$ as follows:

\begin{figure}
    \centering
    \includegraphics[width=0.9\linewidth]{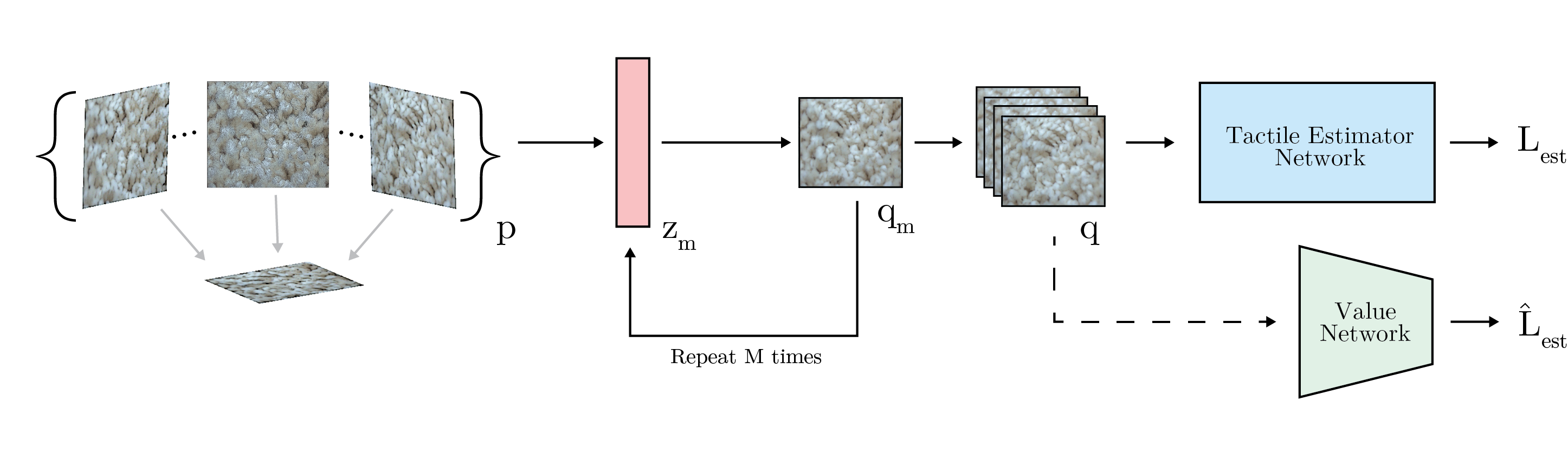}
    \vspace{-0.35cm}
    \caption{\small{\textbf{Viewpoint selection frameworks.} Inspired by recent works in neural architecture search, we construct a network (NVS) to learn which combination $q$ of viewing angles, from all available angles $p$, minimize the tactile estimator loss $\mathrm{L}_{est}$. The NVS network is comprised of $M$ viewpoint selector vectors $z$, each responsible for selecting one viewpoint. An additional value network estimates the loss of the tactile estimator network given the selected viewpoints $q$.}} \vspace{-0.2cm}
    \label{fig:viewpoint-selector-module}
\end{figure}




\begin{equation}
    q_m = \argmax_i \frac{exp(z_{m, i})}{\sum_{n=1}^N exp(z_{m,n})}, m = 1 \dots M.
\end{equation}
The viewpoint selector vector $z$ is defined in $\mathbb{R}^{N}$ space. This process is repeated $M$ times with different viewpoint selector vectors to select a set of $M$ viewpoints. There are no constraints between the vectors, therefore allowing repeated viewpoints. We explore adding constraints to the selected combinations by including a value function $\mathrm{V}(q;w_v)$, which estimates how well the selected combination will perform on tactile property estimation. This network acts as a lightweight proxy for the tactile estimation network. We call this framework the Value Based Neural Viewpoint Selector (VB-NVS). The value function provides additional guidance for the viewpoint selector vectors by propagating an additional error signal from the objective as shown in Figure~\ref{fig:viewpoint-selector-module}. The input of the value function is the combined output probability distribution from the viewpoint selector vectors. The value network then estimates the tactile estimation error $\Est{\mathcal{L}}_{est}$.


Inspired by the gradient-based neural architecture search by Pham et al.~\cite{pham2018efficient}, the weights of the neural viewpoint selector $\pi(w_{\pi})$ and tactile estimator network $f(q;w_\theta)$ are trained independently in three stages. First the weights of the $f(q;w_\theta)$ network are trained with random combinations of $q$. Then all the weights of the tactile estimator network are frozen, excluding the late-fusion layer. Combinations of viewing angles are sampled from the policy network $\pi(w_{\pi})$ and evaluated with $f(q;w_\theta)$. The weights of $\pi(w_{\pi})$ are updated via the gradient produced by the REINFORCE~\cite{williams1992simple} objective. Finally the weights of $f(q;w_\theta)$ are reinitialized and trained with the optimal set of viewing angles $q^\star$ produced by $\pi(w_{\pi})$. The VB-NVS framework is trained in a similar manner except the value network $\mathrm{V}(q;w_v)$ is now trained in conjunction with the viewpoint selector network $\pi(w_{\pi})$ and the optimal set $q^\star$ is generated by the value network instead of $\pi(w_{\pi})$.




\subsubsection{Multi-Image Baseline}

There are a variety of schemes for selecting a subset of data points. Naive approaches include random sampling and sampling at a fixed interval (equidistant). We compare our viewpoint selection networks with these naive approaches along with more advanced algorithms. Zhou et al. subsample image sequences by randomly selecting combinations of various lengths from a given image sequence~\cite{zhou2018temporal}. The random subsamples efficiently find temporal and spatial trends from image sequences outperforming non-sampling approaches for action recognition. Rather than using a subset of viewing angles, we explore using the entire image sequence as input to the tactile property estimator. Su et al. generate feature vectors from each image separately, then fuse each feature vector together via max-pooling~\cite{su2015multi}. Early fusion methods such as I3D use 3D CNNs to learn both image-level features as well as temporal features from concatenated images~\cite{carreira2017quo}. For all multi-image networks except the I3D and View Pooling, we employ late feature fusion with neural networks to aggregate features from multiple images.

\subsubsection{Implementation Details}
The 400+ visuo-tactile pairs in the SPS dataset are randomly divided into 90/10 training/validation splits. Each experiment is rerun three times and the mean score is presented. For the single-image experiments in Section~\ref{section:experiment_single_image}, the image corresponding to the most nadir viewing angle is selected as input. For both single-image (Section~\ref{section:experiment_single_image}) and multi-image (Section~\ref{section:experiment_multi_image}) experiments, the mean of the five tactile measurements is used as the tactile signal. A 50-layered SE-ResNeXt~\cite{hu2018squeeze} network pretrained on ImageNet serves as the image encoding backbone for all networks. We set the size of the latent space to be 50 and 100 for single and multi-image experiments respectively. The networks are trained for 30 epochs with a learning rate of 1e-4. Separate networks are trained for each tactile property. Non-learned and learned sampling methods select combinations of three images ($M=3$). Additional training parameters are described in the supplementary material.

\vspace{-0.4cm}

\begin{table}[]
\centering
\caption{\small{\textbf{Single image tactile estimation ($\mathcal{R}^2$).} The $\mathcal{R}^2$ performance per tactile property is displayed, higher values are better. Our proposed cross-modal model significantly outperforms the baseline regression model across nearly all tactile properties. 
}} \vspace{0.1cm}
\resizebox{1\columnwidth}{!}{%
\begin{tabu}{c|ccccccccccccccc|cc}
\tabucline[1pt]{-}
\addstackgap[3pt]Model & fRS&	cDF&	tCO&	cYD&	aTK&	mTX&	cCM&	cDP&	cRX&	mRG&	mCO&	uRO&	tPR&	uCO&	fST & \multicolumn{1}{|c}{$\mathcal{R}^2$} & \multicolumn{1}{c}{MAE}\\ \hline \hline
Regression & 0.07 & 0.49 & 0.50 & 0.44 & -0.46 & \textbf{0.43} & 0.13 & 0.35 & 0.11 & 0.46 & \textbf{0.56} & 0.32 & 0.57 & 0.57 & 0.53 & 0.34 & 6.17 \\
Cross-Modal & \textbf{0.54} & \textbf{0.52} & \textbf{0.62} & \textbf{0.64} & \textbf{-0.07} & \textbf{0.43} & \textbf{0.47} & \textbf{0.67} & \textbf{0.44} & \textbf{0.47} & 0.54 & \textbf{0.44} & \textbf{0.65} & \textbf{0.59} & \textbf{0.59} & \textbf{0.50} & \textbf{5.53} \\ 
\tabucline[1pt]{-}
\end{tabu}%
}
\label{tab:single-image-tactile-reconstruction}
\end{table}

\vspace{-1.2cm}

\begin{table}[]
\centering
\caption{\small{\textbf{Single image tactile estimation ($\%_{err}$).} The median $\%_{err}$ performance per tactile property is displayed, lower values are better. Tactile properties to the left of the bold center line comprise the top eight percentage error properties.
}} \vspace{0.1cm}
\resizebox{1\columnwidth}{!}{%
\begin{tabu}{c|cccccccc?ccccccc|cc}
\tabucline[1pt]{-}
\addstackgap[3pt]Model & fRS&	cDF&	tCO&	cYD&	aTK&	mTX&	cCM&	cDP&	cRX&	mRG&	mCO&	uRO&	tPR&	uCO&	fST & $\%_{err}$ & $\%_{err}^{T8}$ \\ \hline \hline

Regression & 18.6&	16.6&	18.2&	22.5&	\textbf{12.7}&	28.8&	21.6&	\textbf{21.9}&	34.0&	50.0&	60.4&	65.5&	65.1&	\textbf{70.4}&	80.6 &	39.1 & 20.1\\

Cross-Modal & \textbf{13.0}&	\textbf{15.0}&	\textbf{15.9}&	\textbf{17.2}&	17.3&	\textbf{17.7}&	\textbf{18.9}&	23.4&	\textbf{29.3}&	\textbf{39.3}&	\textbf{49.0}&	\textbf{57.4}&	\textbf{63.3}&	72.0&	\textbf{73.5} & \textbf{34.8} & \textbf{17.3}\\

\tabucline[1pt]{-}
\end{tabu}%
}
\label{tab:single-image-tactile-reconstruction_perr}
\end{table}

\vspace{-0.8cm}

\section{Experiments}
\label{section:experiments}

\subsection{Cross-Modal Experiments}
\label{section:experiment_single_image}

Given a single image of a material surface, our task is to estimate the tactile properties of that surface. 
To highlight the effectiveness of our proposed cross-modal method, we compare the proposed method with a regression network. 
The results of both methods are recorded in Tables~\ref{tab:single-image-tactile-reconstruction} and~\ref{tab:single-image-tactile-reconstruction_perr}. Our proposed single image method outperforms the regression method across almost all fifteen tactile properties achieving better average $\mathcal{R}^2$, MAE, $\%_{err}$, and $\%_{err}^{T8}$ scores. Both networks achieve negative $\mathcal{R}^2$ scores for the adhesive tack (aTK) dimension, hence the estimates for this dimension are worse than using the average training value as a prediction. 
In general, the problem of estimating direct tactile properties from images-only is challenging and we expect a non-trivial margin of error.


\vspace{-0.7cm}

\begin{table}[]
\centering
\caption{\small{\textbf{Single image cross-modal ablation.} Refactoring the network as a cross-modal network with an adversarial objective greatly improves estimation performance. Visuo-tactile cluster labels outperform both material and tactile cluster labels.}} \vspace{0.1cm}
\resizebox{0.95\columnwidth}{!}{%
\begin{tabu}{ccccc|cccc}
\tabucline[1pt]{-}
\multicolumn{1}{c|}{\multirow{2}{*}{Cross-Modal}} & \multicolumn{1}{c|}{\multirow{2}{*}{Adversarial}} & \multicolumn{1}{c|}{Material} & \multicolumn{1}{c|}{Tactile Cluster} & Visuo-Tactile & \multicolumn{4}{c}{Metrics} \\ 
\multicolumn{1}{c|}{} & \multicolumn{1}{c|}{} & \multicolumn{1}{c|}{Classification} & \multicolumn{1}{c|}{Classification~\cite{yuan2017connecting}} & Classification & $\mathcal{R}^2$ & MAE & $\%_{err}$ & $\%_{err}^{T8}$ \\ \hline \hline
 &  &  &  &  & 0.34 & 6.17 & 39.1 & 20.1 \\
\checkmark &  &  &  &  & 0.46 & 5.65 & 34.3 & 17.1 \\
\checkmark & \checkmark &  &  &  & 0.49 & 5.61 & 36.2 & 18.5 \\
\checkmark &  & \checkmark & & & 0.45 & 5.73  & 36.9 & 19.0\\
\checkmark &  &  & \checkmark & & 0.48 & 5.60 & 35.7 & 17.7\\
\checkmark &  &  &  & \checkmark & 0.49 & 5.58 & \textbf{33.8} & \textbf{16.9}\\
\checkmark & \checkmark &  &  & \checkmark & \textbf{0.50} & \textbf{5.53} & 34.8 & 17.3\\
\tabucline[1pt]{-}
\end{tabu}
}
\label{tab:single-image-cross-modal-ablation}
\end{table}


In order to access the contribution of each component of the cross-modal network, we conduct an ablation study. In Table~\ref{tab:single-image-cross-modal-ablation}, the performance of the baseline regression network is compared to cross-modal networks with auxiliary objectives. We additionally explore using various auxiliary classification label types. As shown in Table~\ref{tab:single-image-cross-modal-ablation}, refactoring the network as a cross-modal network significantly improves the performance of the tactile estimation from an average $\mathcal{R}^2$/$\%_{err}$ of 0.34/39.1 to 0.46/34.3. 
Next, the contribution of the adversarial objective is assessed and we find that the conditional GAN objective improves the quality of generated tactile samples in terms of average $\mathcal{R}^2$ and MAE but not $\%_{err}$ or $\%_{err}^{T8}$. We then evaluate the performance of using different labels for the auxiliary latent space classification task. Using the material class labels, shown in Figure~\ref{fig:dataset_stats}a, degrades the overall performance of the network. This suggests that traditional material labels do not adequately represent the tactile properties of a surface. 
The tactile cluster labels~\cite{yuan2017connecting} improve results but not as much as our proposed joint visuo-tactile labels. 

\subsection{Viewing Angle Selection Experiments}
\label{section:experiment_multi_image}


After examining various network modules and frameworks for tactile property estimation from a single image, we investigate utilizing multiple images as input to the system. As described in Section~\ref{sec:methods}, there are many ways of selecting a subset of images including non-learning methods such as random, equidistant, or TRN sampling and learned methods such as NVS and VB-NVS. In Tables~\ref{tab:multi-image-tactile-reconstruction} and~\ref{tab:multi-image-tactile-reconstruction_perr}, we compare various viewpoint sampling methods. 
Our proposed NVS and VB-NVS methods outperform all other multi-image methods in terms of average $\mathcal{R}^2$, MAE, $\%_{err}$, and $\%_{err}^{T8}$. Surprisingly, all methods that utilize the full amount of available imagery, i.e. Late Fusion, I3DNet~\cite{carreira2017quo}, and View Pooling~\cite{su2015multi}, perform much worse than the single image methods. The poor performance of the I3DNet architecture is likely a consequence of lack of significant inter-frame change in our image sequences. 
The View Pooling method slightly outperforms the other late fusion method. Non-learning sampling methods such as random sampling, equidistant sampling, and TRN~\cite{zhou2018temporal} select subsamples from the total set of viewing angles without updating the selection based on performance.
The non-learned sampling methods surpass the performance of the single image model on several of the tactile properties with only equidistant sampling outperforming the single image method on average. 
Both NVS and VB-NVS achieve the best performance on average across all metrics while providing insightful viewpoint selection information. However, they do not outperform the single image methods in several categories suggesting that multiple images do not always provide useful information for estimating certain tactile properties. None of the multi-image methods are able to consistently provide a better than average prediction for the adhesive tack property (aTK).

\vspace{-0.75cm}

\begin{table}[]
\centering
\caption{\small{\textbf{Multi-Image tactile estimation ($\mathcal{R}^2$).} The $\mathcal{R}^2$ performance per tactile property is displayed, higher values are better. The proposed viewpoint selection frameworks outperform all other models on average. {\color{red} Red} and {\color{blue} blue} text correspond to the first and second best scores respectively.
}
} \vspace{0.1cm}
\resizebox{1\columnwidth}{!}{%
\begin{tabu}{c|ccccccccccccccc|cc}
\tabucline[1pt]{-}
\addstackgap[3pt]Model & fRS&	cDF&	tCO&	cYD&	aTK&	mTX&	cCM&	cDP&	cRX&	mRG&	mCO&	uRO&	tPR&	uCO&	fST & \multicolumn{1}{|c}{$\mathcal{R}_2$} & \multicolumn{1}{c}{MAE}\\ \hline \hline

Single (\TextRed{ours}) & 0.54 & {\color{blue}0.52} &  {\color{blue}0.62} & {\color{red}0.64} & -0.07 & 0.43 & 0.47 & {\color{red}0.67} & 0.44 & 0.47 & 0.54 & 0.44 & {\color{blue}0.65} & 0.59 & 0.59 & 0.50 & 5.53\\ \hline

Late-Fusion & 
0.05 & -0.06 & 0.04 & -0.21 & -0.15 & 0.46 & 0.37 & 0.04 & -0.10 & 0.31 & 0.35 & 0.04 & 0.33 & -0.01 & 0.23 & 0.11 & 9.16 \\

I3DNet~\cite{carreira2017quo} & 
0.32 & 0.01 & -0.23 & 0.04 & -0.23 & 0.37 & 0.31 & 0.12 & -0.13 & 0.24 & 0.11 & 0.07 & 0.09 & -0.16 & 0.14 & 0.11 & 9.42 \\

View Pooling~\cite{su2015multi} & 
0.03 & 0.03 & -0.01 & -0.09 & -0.11 & 0.47 & 0.23 & 0.07 & -0.05 & 0.30 & 0.32 & 0.12 & 0.25 & -0.02 & 0.10 & 0.11 & 9.01 \\ \hline

Random & 
0.49 & 0.41 & 0.58 & 0.49 & {\color{blue}-0.02} & 0.45 & 0.46 & 0.57 & 0.31 & 0.46 & {\color{red}0.72} & 0.40 & 0.56 & 0.55 & 0.59 & 0.47 & 5.49 \\ 

Equidistant & 
0.63 & 0.48 & {\color{blue}0.62} & 0.50 & -0.04 & 0.52 & 0.52 & 0.58 & 0.41 & 0.45 & 0.61 & 0.44 & 0.61 & 0.62 & {\color{red}0.65} & 0.51 & 5.37 \\ 

TRN~\cite{zhou2018temporal} & 
{\color{blue}0.64} & 0.49 & 0.52 & 0.40 & -0.10 & 0.52 & {\color{blue}0.55} & 0.61 & 0.35 & 0.41 & 0.66 & 0.39 & 0.53 & 0.50 & 0.63 & 0.47 & 5.53 \\ \hline

NVS (\TextRed{ours}) & 
0.62 & {\color{red}0.53} & {\color{red}0.63} & 0.55 & {\color{red}0.02} & {\color{blue}0.56} & 0.53 & 0.54 & {\color{red}0.49} & {\color{blue}0.55} & 0.68 & {\color{red}0.51} & 0.54 & {\color{blue}0.63} & {\color{blue}0.64} & {\color{blue}0.53} & {\color{blue}5.34} \\ 

VB-NVS (\TextRed{ours}) & 
{\color{red}0.65} & 0.50 & 0.61 & {\color{blue}0.57} & -0.05 & {\color{red}0.58} & {\color{red}0.57} & {\color{blue}0.64} & {\color{blue}0.45} & {\color{red}0.57} & {\color{blue}0.70} & {\color{blue}0.47} & {\color{red}0.68} & {\color{red}0.66} & 0.61 & {\color{red}0.55} & {\color{red}5.28} \\ 

\tabucline[1pt]{-}
\end{tabu}%
}
\label{tab:multi-image-tactile-reconstruction}
\end{table}

\vspace{-1.3cm}

\begin{table}[]
\centering
\caption{\textbf{Multi-Image tactile estimation ($\%_{err}$).} \small{The median $\%_{err}$ performance per tactile property is displayed, lower values are better. The proposed viewpoint selection frameworks outperform all other models on average. {\color{red} Red} and {\color{blue} blue} text correspond to the first and second best scores respectively.}
} \vspace{0.1cm}
\resizebox{1\columnwidth}{!}{%
\begin{tabu}{c|ccccccccccccccc|cc}
\tabucline[1pt]{-}
\addstackgap[3pt]Model & fRS&	cDF&	tCO&	cYD&	aTK&	mTX&	cCM&	cDP&	cRX&	mRG&	mCO&	uRO&	tPR&	uCO&	fST & $\%_{err}$ & $\%_{err}^{T8}$ \\ \hline \hline
Single & {\color{red}13.0} & {\color{blue}15.0} & 15.9 & 17.2 & 17.3 & 17.7 & 18.9 & 23.4 & {\color{red}29.3} & 39.3 & {\color{red}49.0} & {\color{red}57.4} & 63.3 & 72.0 & 73.5 & 34.8 & 17.3 \\ \hline

Late-Fusion & 
30.4 & 18.6 & 24.6 & 35.6 & 14.3 & 28.5 & 20.1 & 25.9 & 36.7 & {\color{blue}32.6} & 244.7 & 251.6 & 91.1 & 74.3 & 78.1 & 67.1 & 24.7 \\

I3DNet~\cite{carreira2017quo} & 
29.9 & 19.9 & 29.4 & 25.5 & 13.9 & 24.1 & 30.6 & 36.2 & 32.7 & 54.3 & 231.1 & 131.4 & 77.7 & 85.4 & 74.2 & 59.8 & 26.2 \\ 

View Pooling~\cite{su2015multi} & 
50.4 & 30.9 & 22.8 & 34.7 & 19.5 & 18.7 & 48.6 & 34.4 & 35.1 & 44.6 & 97.2 & 64.4 & 174.6 & 169.0 & 140.3 & 65.7 & 32.5 \\ \hline

Random & 
{\color{blue}13.2} & 18.1 & 13.7 & 17.0 & 13.4 & 15.4 & 19.2 & {\color{red}22.8} & 34.7 & 38.2 & 54.7 & 62.2 & 59.3 & {\color{red}59.6} & {\color{red}50.8} & 32.8 & 16.6 \\ 

Equidistant & 
16.3 & 17.2 & 12.9 & 20.7 & {\color{red}9.4} & {\color{blue}12.9} & 19.8 & {\color{blue}23.3} & 35.0 & 34.8 & 52.3 & 63.0 & 60.2 & {\color{blue}62.4} & 58.5 & 33.2 & 16.6 \\ 

TRN~\cite{zhou2018temporal} & 
16.9 & 16.1 & {\color{blue}12.4} & 14.8 & 11.7 & 13.7 & 23.4 & 25.2 & 35.3 & 37.4 & 56.1 & 63.2 & 60.1 & 63.6 & 54.0 & 33.6 & 16.8 \\ \hline

NVS (\TextRed{ours}) & 
15.0 & 15.4 & 12.8 & {\color{blue}13.7} & 10.6 & {\color{red}11.8} & {\color{blue}18.3} & 24.5 & 33.3 & 37.8 & {\color{blue}51.2} & 64.2 & {\color{blue}56.3} & 62.9 & 58.3 & {\color{blue}32.4} & {\color{red}15.3} \\ 

VB-NVS (\TextRed{ours}) & 
16.3 & {\color{red}13.6} & {\color{red}11.5} & {\color{red}12.4} & {\color{blue}9.6} & 17.2 & {\color{red}17.6} & 26.8 & {\color{blue}31.8} & {\color{red}32.5} & 53.2 & {\color{blue}59.8} & {\color{red}55.1} & 67.3 & {\color{blue}53.9} & {\color{red}31.9} & {\color{blue}15.6} \\ 

\tabucline[1pt]{-}
\end{tabu}%
}
\label{tab:multi-image-tactile-reconstruction_perr}
\end{table}

\vspace{-0.5cm}

\begin{figure}
    \centering
    \includegraphics[width=0.9\linewidth, height=0.35\columnwidth]{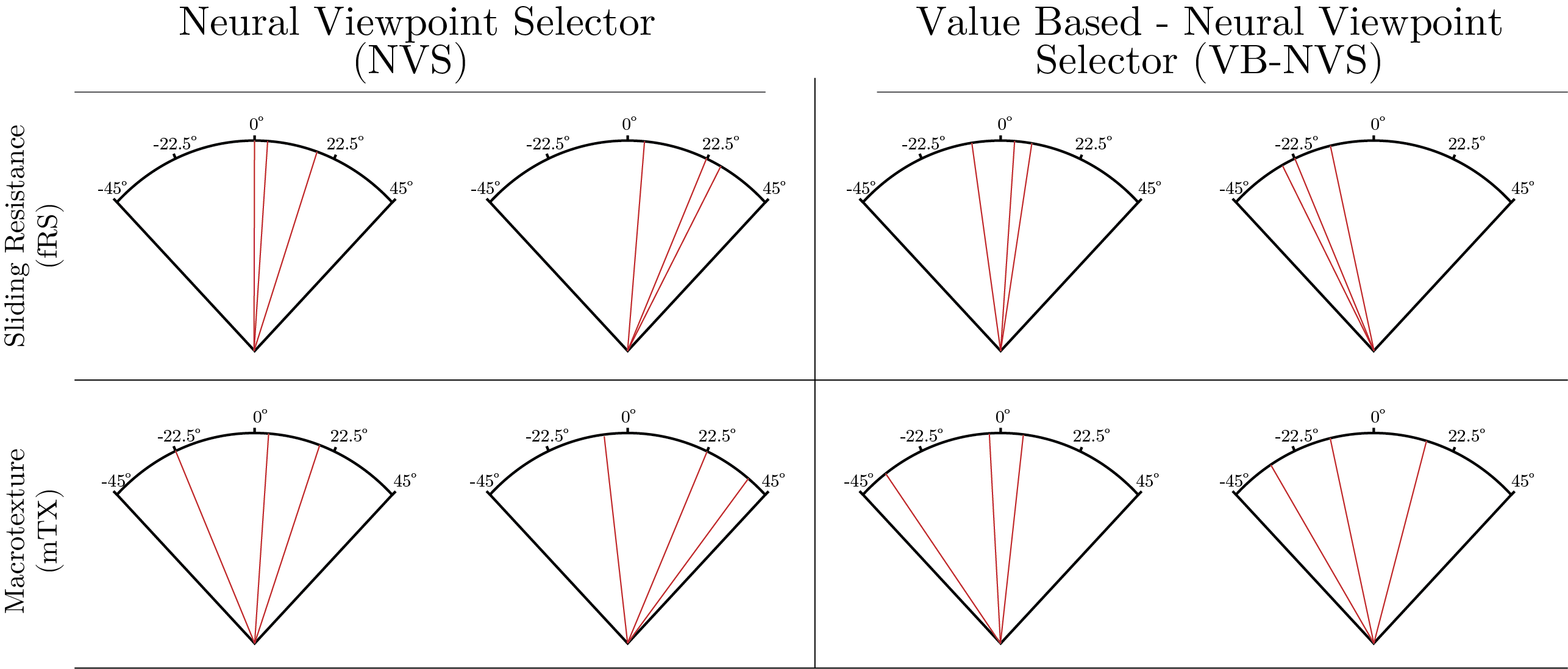}
    \caption{\small{\textbf{Viewpoint selection result.} The resultant selected viewpoints of both learned sampling methods. Columns represent repeated experiments, highlighting the consistency of the selected viewing angle combinations. Models optimized to estimate the sliding resistance property learn to select viewpoints that are close in viewing angle space while the selected viewpoints for the macrotexture property are farther apart.}
    }
    \label{fig:viewing_angle_selection}
\end{figure}

In addition to improved performance from the learned subsampling methods, we gain insight into which combinations of viewing angles are useful for estimating a specific physical property. In Figure~\ref{fig:viewing_angle_selection}, the selected viewing angles from trained NVS and VB-NVS modules are shown for several tactile properties. Note, this visualization is per tactile property, not per material. The rows of Figure~\ref{fig:viewing_angle_selection} represent the viewing angles selected for a particular tactile property while the columns represent repeated experiments.  Selected viewpoints for models trained to estimate sliding resistance (fRS) are consistently close in viewing angle space for both the NVS and VB-NVS methods. The distribution of viewing angles does not vary considerably across each experiment but the location of the distribution does. This suggests that the relative difference between viewing angles is more important for our objective than the global values of the viewing angles.
The viewing angle selection is consistent with observations of prior work that angular gradients are important for material recognition \cite{zhang2015reflectance,xue2017differential,wang20164d}.
Similar trends are observed for the macrotexture (mTX) viewing angle subsamples. The difference between the selected viewing angles for the mTX property is greater than those of the fST property suggesting that wider viewing angles are preferable to estimate macrotexture properties. 

\vspace{-0.25cm}

\vspace{-1.0cm}

\section{Conclusion}

This work is a pioneering step towards understanding the relationship between visual and tactile information. We propose a new challenge of estimating fifteen tactile physical properties of a surface from multiview images. We provide several methods that estimate tactile properties and determine the optimal viewing angles to sample for the estimation. To train our models we assemble the first of its kind, visuo-tactile dataset containing tactile physical properties and corresponding image sequences. We tackle the challenge of physical property estimation by designing a cross-modal network with an adversarial and a joint classification objective with results that surpass prior work in cross-modal translation.  Additionally, our viewpoint selection framework achieves state-of-the-art performance for this task while providing insight as to which  combinations of viewing angles are optimal for estimating a given tactile property. The proposed method can be used directly or as a prior for several tasks such as  automated driving (road condition estimation), robotics (object manipulation or navigation) and manufacturing (quality control).

\paragraph{Acknowledgments} This research was supported by NSF Grant \#1715195. We would like to thank Eric Wengrowski, Peri Akiva, and Faith Johnson for the useful suggestions and discussions.



%
%

\bibliographystyle{splncs04}
\bibliography{egbib}
\end{document}